\documentclass[12pt,letterpaper]{article}
\usepackage[a4paper, total={7in, 10in}]{geometry}

\usepackage{graphicx}
\usepackage{helvet}
\usepackage{authblk}
\usepackage{hyperref}
\usepackage{amsmath} 
\usepackage{amssymb} 
\usepackage{orcidlink} 
\usepackage[super,comma,sort&compress]  
   {natbib}
\usepackage[right]{lineno} \linenumbers
\usepackage[title]{appendix}%

\usepackage[utf8]{inputenc} 
\usepackage[T1]{fontenc}    
\usepackage{url}            
\usepackage{booktabs}       
\usepackage{amsfonts}       
\usepackage{nicefrac}       
\usepackage{microtype}      
\usepackage{xcolor}         
\usepackage{multirow}
\usepackage{threeparttable}
\usepackage{amsmath}
\usepackage{subcaption}
\usepackage{makecell}
\usepackage{placeins}
\usepackage{booktabs}      
\usepackage{siunitx}       
\usepackage{caption}       
\usepackage{tabularx}
\usepackage{listings}    
\usepackage{pifont}
\usepackage{array}
\usepackage{makecell}
\theadset{\bfseries}

\makeatletter
\renewcommand{\maketitle}{\bgroup\setlength{\parindent}{0pt}
\begin{flushleft}
  \textbf{\@title}
  
  \@author
\end{flushleft}\egroup}
\makeatother

\title{Artificial intelligence-enabled single-lead ECG for non-invasive hyperkalemia detection: development, multicenter validation, and proof-of-concept deployment}

\date{}

\author[1,2,\#,\orcidlink{0009-0007-9505-3699}]{Gongzheng Tang}
\author[3,\#]{Qinghao Zhao}
\author[4,\#]{Guangkun Nie}
\author[1,2]{Yujie Xiao}
\author[5]{Shijia Geng}
\author[1,2]{Donglin Xie}
\author[2]{Shun Huang}
\author[5]{Deyun Zhang}
\author[6,7,8]{Xingchen Yao}
\author[6,7,8]{Jinwei Wang}
\author[9,*]{Kangyin Chen}
\author[2,6,7,8,*]{Luxia Zhang}
\author[1,2,10,11,*,\orcidlink{0000-0001-7521-5127}]{Shenda Hong}

\affil[1]{Institute of Medical Technology, Peking University Health Science Center, Beijing, China}
\affil[2]{National Institute of Health Data Science, Peking University, Beijing, China}
\affil[3]{Department of Cardiology, Peking University People’s Hospital, Beijing, China}
\affil[4]{School of Intelligence Science and Technology, Peking University, Beijing, China}
\affil[5]{Heart Voice Medical Technology, Hefei, China}
\affil[6]{Renal Division, Department of Medicine, Peking University First Hospital, Beijing, China}
\affil[7]{Institute of Nephrology, Key Laboratory of Renal Disease, Ministry of Health of China, and Key Laboratory of Chronic Kidney Disease Prevention and Treatment (Peking University), Ministry of Education, Beijing, China}
\affil[8]{Research Units of Diagnosis and Treatment of Immune-mediated Kidney Diseases, Chinese Academy of Medical Sciences, Beijing, China}
\affil[9]{Tianjin Key Laboratory of Ionic-Molecular Function of Cardiovascular Disease, Department of Cardiology, Tianjin Institute of Cardiology, The Second Hospital of Tianjin Medical University, Tianjin, China}
\affil[10]{State Key Laboratory of Vascular Homeostasis and Remodeling, NHC Key Laboratory of Cardiovascular Molecular Biology and Regulatory Peptides, Peking University, Beijing, China} 
\affil[11]{Institute for Artificial Intelligence, Peking University, Beijing, China}

\affil[$\#$]{These authors contributed equally}
\affil[*]{Correspondence: hongshenda@pku.edu.cn, zhanglx@bjmu.edu.cn, chenkangyin@vip.126.com}

\begin{document}

\maketitle

\section*{ABSTRACT}

Hyperkalemia is a life-threatening electrolyte disorder that is common in patients with chronic kidney disease and heart failure, yet frequent monitoring remains difficult outside hospital settings. We developed and validated Pocket-K, a single-lead AI-ECG system initialized from the ECGFounder foundation model for non-invasive hyperkalemia screening and handheld deployment. In this multicentre observational study using routinely collected clinical ECG and laboratory data, 34,439 patients contributed 62,290 ECG--potassium pairs. Lead I data were used to fine-tune the model. Data from Peking University People's Hospital were divided into development and temporal validation sets, and data from The Second Hospital of Tianjin Medical University served as an independent external validation set. Hyperkalemia was defined as venous serum potassium $> 5.5$ mmol/L. Pocket-K achieved AUROCs of 0.936 in internal testing, 0.858 in temporal validation, and 0.808 in external validation. For KDIGO-defined moderate-to-severe hyperkalemia (serum potassium $\geq$ 6.0 mmol/L), AUROCs increased to 0.940 and 0.861 in the temporal and external sets, respectively. External negative predictive value exceeded 99.3\%. Model-predicted high risk below the hyperkalemia threshold was more common in patients with chronic kidney disease and heart failure. A handheld prototype enabled near-real-time inference, supporting future prospective evaluation in native handheld and wearable settings.


\section*{INTRODUCTION}

Hyperkalemia, defined as a venous serum potassium concentration above 5.5 mmol/L, is a common and potentially fatal electrolyte abnormality and a major risk factor for malignant arrhythmia and sudden cardiac death \cite{kovesdy2016epidemiology}. It is particularly common in patients with chronic kidney disease and heart failure, and its incidence increases substantially as renal function declines \cite{hougen2021hyperkalemia,mclean2022population}. Previous studies have reported a U-shaped association between potassium concentration and mortality, with hyperkalemia linked to a higher risk of death in chronic kidney disease, heart failure, and acute myocardial infarction \cite{kovesdy2018serum,goyal2012serum,lin2026ai}. Timely intervention therefore depends on rapid and reliable recognition.

Confirmation of hyperkalemia currently relies on venous blood sampling and laboratory measurement of potassium. Although this standard is accurate, it is invasive, intermittent, and dependent on access to clinical facilities, limiting frequent monitoring between routine visits for high-risk patients \cite{asirvatham2013errors}. Hyperkalemia also produces characteristic electrocardiographic changes, including peaked T waves and QRS widening, which makes the electrocardiogram a plausible non-invasive screening tool \cite{campese2016electrophysiological}. However, visual interpretation alone has limited sensitivity, even among experienced clinicians, which reduces the reliability of the electrocardiogram for screening \cite{lin2022point} and leaves hyperkalemia management an ongoing clinical challenge.

Artificial intelligence (AI), particularly deep learning, has been increasingly applied in clinical research and cardiovascular medicine \cite{ai4s1,ai4s2,ai4s3,ai4s4,ai4s5,ai4s6,ai4s7}. In electrocardiography, AI-based methods can detect subtle waveform features that are difficult to recognise by routine visual inspection and have shown promise in identifying cardiovascular and metabolic abnormalities \cite{research1,research2,research3,research4,research5}. In hyperkalemia, models trained on standard 12-lead ECGs have shown strong diagnostic performance and can capture risk signals that are subclinical or missed by conventional interpretation \cite{lin2020deep,harmon2024mortality,chen2025monitoring}. However, the workflow required to acquire a 12-lead ECG limits its use for repeated surveillance and self-screening outside the hospital, where scalable monitoring may be most valuable for high-risk patients. By contrast, single-lead ECG, particularly Lead I, is the dominant format used by handheld recorders and smartwatches, making it well suited to home monitoring and out-of-hospital screening. Small single-centre studies from the USA, Taiwan, and other settings have provided early evidence that single-lead hyperkalemia detection is feasible \cite{attia2016novel,urtnasan2022noninvasive,chiu2024serum}. However, the current evidence base remains limited by modest sample size, restricted population diversity, and the absence of rigorous large-scale multicentre validation. Consequently, performance across heterogeneous populations, different health systems and acquisition platforms, and real-world prevalence settings remains uncertain.

We therefore developed Pocket-K, a single-lead AI-ECG model for hyperkalemia screening, initialised from a pretrained ECG foundation model. We evaluated model performance in three settings: internal testing within the development set, temporally separated validation within the same health system, and independent external validation in a second health system using a different ECG platform. We also examined waveform-level interpretability, patient-level longitudinal trajectories, clinical phenotypes associated with false-positive predictions, and the technical feasibility of handheld deployment.

\section*{RESULTS}

\subsection*{Study Overview}

\begin{figure}[htbp]
    \centering
    \includegraphics[width=\textwidth]{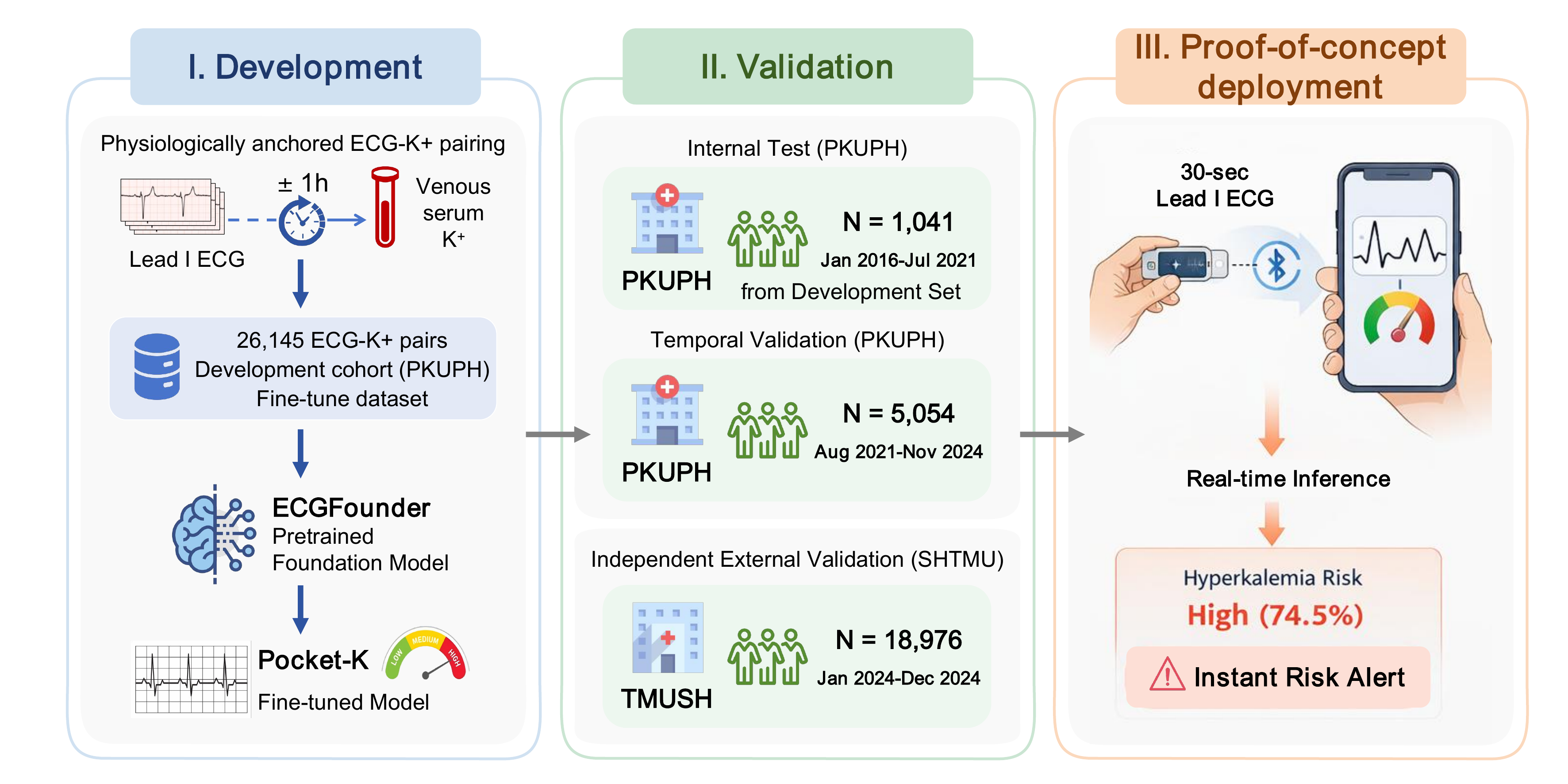}
    \caption{Overview of Pocket-K development, validation, and proof-of-concept deployment. In the development stage, ECG--$K^+$ pairs were constructed by linking each ECG to the nearest eligible venous serum potassium measurement within a $\pm$1-hour window. These paired data were used to fine-tune ECGFounder, a pretrained ECG foundation model, to develop Pocket-K. In the validation stage, model performance was assessed in three settings: an internal test set from the PKUPH development set, a temporal validation set at PKUPH, and an independent external validation set from SHTMU. In the proof-of-concept stage, a handheld device recorded a 30-s lead I ECG, which was processed through a smartphone-based workflow for near-real-time inference and generation of a hyperkalemia risk alert.}
    \label{fig:overview}
\end{figure}

We developed Pocket-K, a single-lead ECG model for non-invasive hyperkalemia screening that was initialized from a pretrained ECG foundation model. The overall study design is shown in Figure~\ref{fig:overview}. The study used a multistage design. We first constructed ECG--$K^+$ pairs by linking each ECG to a venous serum potassium measurement obtained within a $\pm$1-hour window. We then fine-tuned the pretrained foundation model, ECGFounder, to estimate the probability of hyperkalemia, defined as serum $K^+ > 5.5$ mmol/L.

We evaluated the model in 34,439 unique patients from two independent health systems. At the primary site, Peking University People's Hospital(PKUPH), patients were divided chronologically at July 2021 to examine performance over time, yielding a development set ($N=10{,}409$; $n=26{,}145$ pairs) and a temporal validation set ($N=5{,}054$; $n=10{,}922$ pairs). We then tested the model in an independent external validation set from The Second Hospital of Tianjin Medical University (SHTMU) ($N=18{,}976$; $n=25{,}223$ pairs), which included different ECG management systems, hardware environments, and care settings.

We also conducted a proof-of-concept evaluation to assess whether the model could support near-real-time, out-of-hospital screening in high-risk populations using handheld single-lead devices. This study design allowed us to evaluate Pocket-K both in historical clinical datasets and in a practical prototype deployment setting.

\subsection*{Study population and baseline characteristics}
The final study population included 34,439 unique patients and 62,290 ECG--$K^+$ pairs from two health systems. After quality control at PKUPH, 15,463 patients contributed 37,067 eligible pairs. Of these, 10,409 patients with 26,145 pairs formed the development set, and 5,054 patients with 10,922 pairs formed the temporal validation set after exclusion of patients who overlapped with the development set. SHTMU contributed 18,976 patients with 25,223 pairs for independent external validation.

Baseline clinical characteristics and technical specifications are summarised in Table~\ref{tab:baseline}. Across sets, mean age ranged from 54.08 to 58.25 years, and men accounted for 53.3\%--56.0\% of patients. Compared with the development set, the external validation set had a higher burden of renal dysfunction and cardiorenal comorbidity. In all sets, the mean interval between ECG acquisition and potassium measurement was less than 30 min, supporting close physiological alignment between waveform acquisition and the laboratory reference.

\begin{table}[htbp]
\centering
\caption{Baseline clinical characteristics, ECG--potassium pairing quality, and technical specifications across study sets. Continuous variables are presented as mean (SD), and categorical variables as n (\%). PKUPH: Peking University People's Hospital; SHTMU: The Second Hospital of Tianjin Medical University; $K^+$: serum potassium; eGFR: estimated glomerular filtration rate; ECG: electrocardiogram.}
\label{tab:baseline}

\footnotesize
\setlength{\tabcolsep}{4pt}
\renewcommand{\arraystretch}{1.18}

\begin{threeparttable}
\begin{tabularx}{\textwidth}{l>{\centering\arraybackslash}X>{\centering\arraybackslash}X>{\centering\arraybackslash}X}
\toprule
\textbf{Characteristic} &
\textbf{Development Set \  (PKUPH)} &
\textbf{Temporal Validation Set \  (PKUPH)} &
\textbf{Independent External Validation Set \  (SHTMU)} \\
&
\textbf{($N=10{,}409$; $n=26{,}145$)} &
\textbf{($N=5{,}054$; $n=10{,}922$)} &
\textbf{($N=18{,}976$; $n=25{,}223$)} \\
\midrule

\multicolumn{4}{l}{\textbf{Demographics}} \\
Age, years & 54.08 (15.55) & 54.60 (14.70) & 58.25 (17.90) \\
Male sex, n (\%) & 5,642 (54.2\%) & 2,830 (56.0\%) & 9,994 (53.3\%) \\

\addlinespace[2pt]

\multicolumn{4}{l}{\textbf{Laboratory measurements}} \\
Serum $K^+$, mmol/L & 4.14 (0.36) & 4.19 (0.40) & 4.11 (0.42) \\
Serum creatinine, $\mu$mol/L & 74.66 (95.29) & 78.68 (173.15) & 106.08 (380.26) \\
eGFR, mL/min/1.73\,m$^2$ & 95.80 (23.14) & 93.84 (21.91) & 93.92 (28.84) \\
\addlinespace[2pt]

\multicolumn{4}{l}{\textbf{eGFR-based categories, n (\%)}} \\
$\geq$90 & 8,757 (84.1\%) & 3,915 (77.5\%) & 12,523 (66.0\%) \\
60--89 & 1,291 (12.4\%) & 932 (18.4\%) & 4,418 (23.3\%) \\
30--59 & 242 (2.3\%) & 143 (2.8\%) & 1,241 (6.5\%) \\
$<$30 & 119 (1.1\%) & 64 (1.3\%) & 794 (4.2\%) \\
\addlinespace[2pt]

\multicolumn{4}{l}{\textbf{ECG--potassium pairing quality}} \\
ECG-to-$K^+$ interval, min & 28.46 (14.88) & 24.44 (14.41) & 22.39 (16.26) \\
\addlinespace[2pt]

\multicolumn{4}{l}{\textbf{Comorbidities, n (\%)}} \\
Hypertension & 2,263 (21.7\%) & 1,453 (28.7\%) & 6,778 (35.7\%) \\
Diabetes mellitus & 1,041 (10.0\%) & 690 (13.7\%) & 3,137 (16.5\%) \\
Chronic kidney disease & 430 (4.1\%) & 238 (4.7\%) & 2,127 (11.2\%) \\
Heart failure & 68 (0.7\%) & 57 (1.1\%) & 437 (2.3\%) \\
Coronary artery disease & 765 (7.3\%) & 556 (11.0\%) & 2,463 (13.1\%) \\
Stroke & 444 (4.3\%) & 262 (5.2\%) & 1,357 (7.2\%) \\
\addlinespace[2pt]

\multicolumn{4}{l}{\textbf{Technical specifications}} \\
ECG management system & MedEx & MedEx & Nalong \\
Dominant hardware & GE / Philips & GE / Philips & Nalong / Nihon Kohden \\
Sampling frequency, Hz & 500 & 500 & 500 / 1000 \\
\bottomrule
\end{tabularx}

\begin{tablenotes}[flushleft]
\footnotesize
\item $N$ denotes the number of unique patients, and $n$ denotes the number of ECG--potassium pairs.
\item Technical specifications are reported at the set level rather than the patient level.
\end{tablenotes}
\end{threeparttable}
\end{table}

\subsection*{Pocket-K generalized across internal, temporal, and independent external validation sets}

\begin{figure}[htbp]
    \centering
    \captionsetup{font=small, labelfont=bf}

    \begin{subfigure}{0.32\textwidth}
        \centering
        \includegraphics[width=\linewidth]{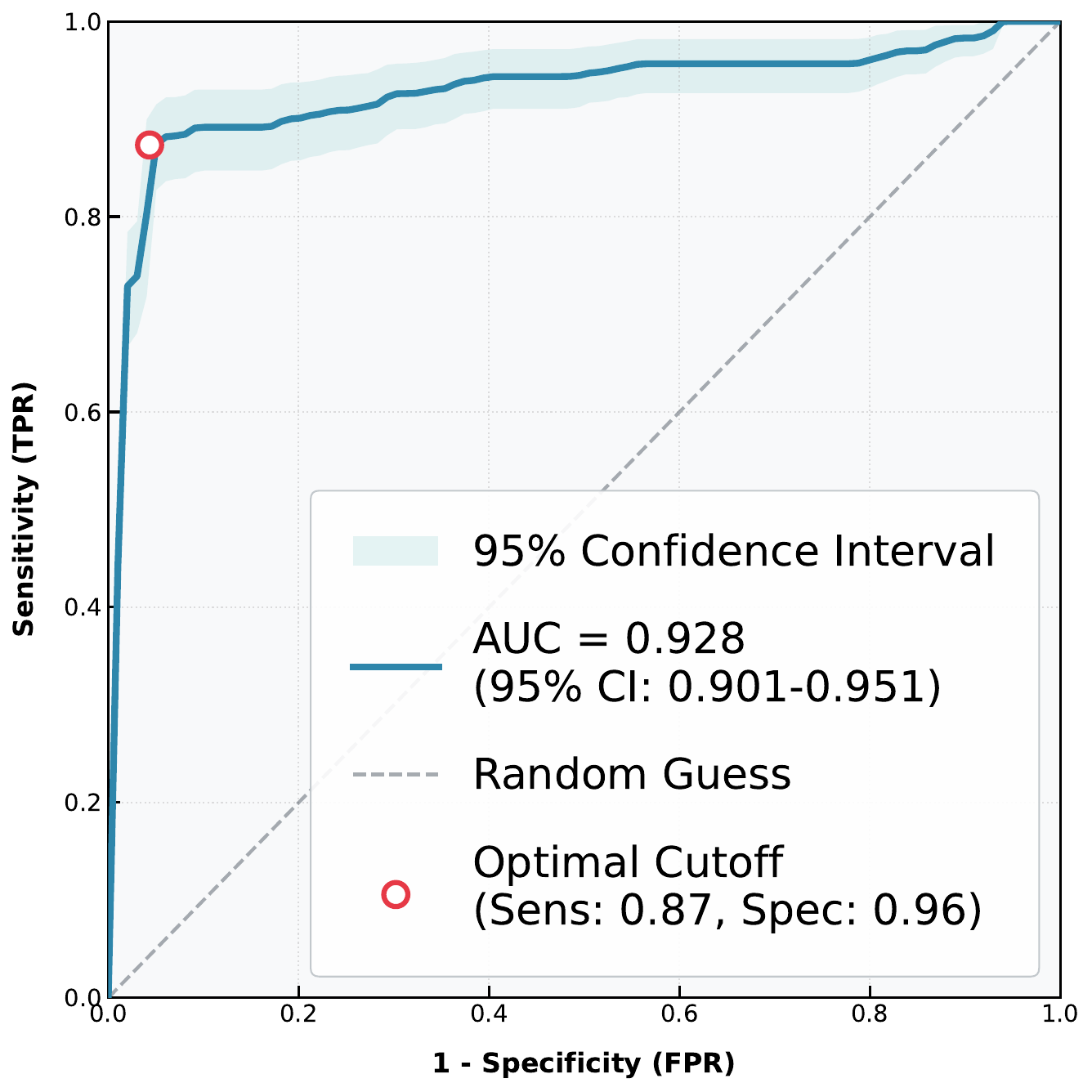}
        \caption{Internal Test}
        \label{fig:results_roc_a}
    \end{subfigure}
    \hfill
    \begin{subfigure}{0.32\textwidth}
        \centering
        \includegraphics[width=\linewidth]{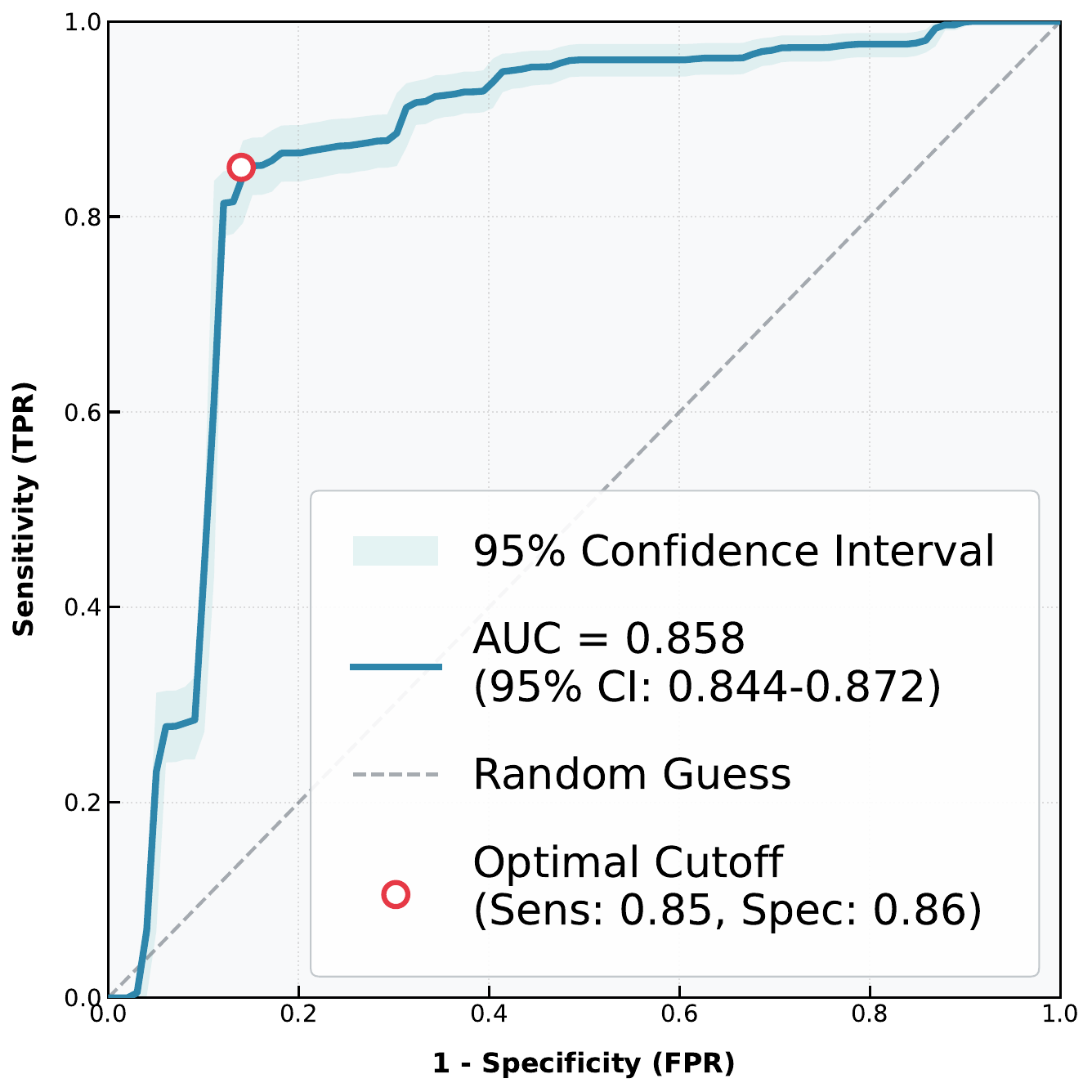}
        \caption{Temporal validation}
        \label{fig:results_roc_b}
    \end{subfigure}
    \hfill
    \begin{subfigure}{0.32\textwidth}
        \centering
        \includegraphics[width=\linewidth]{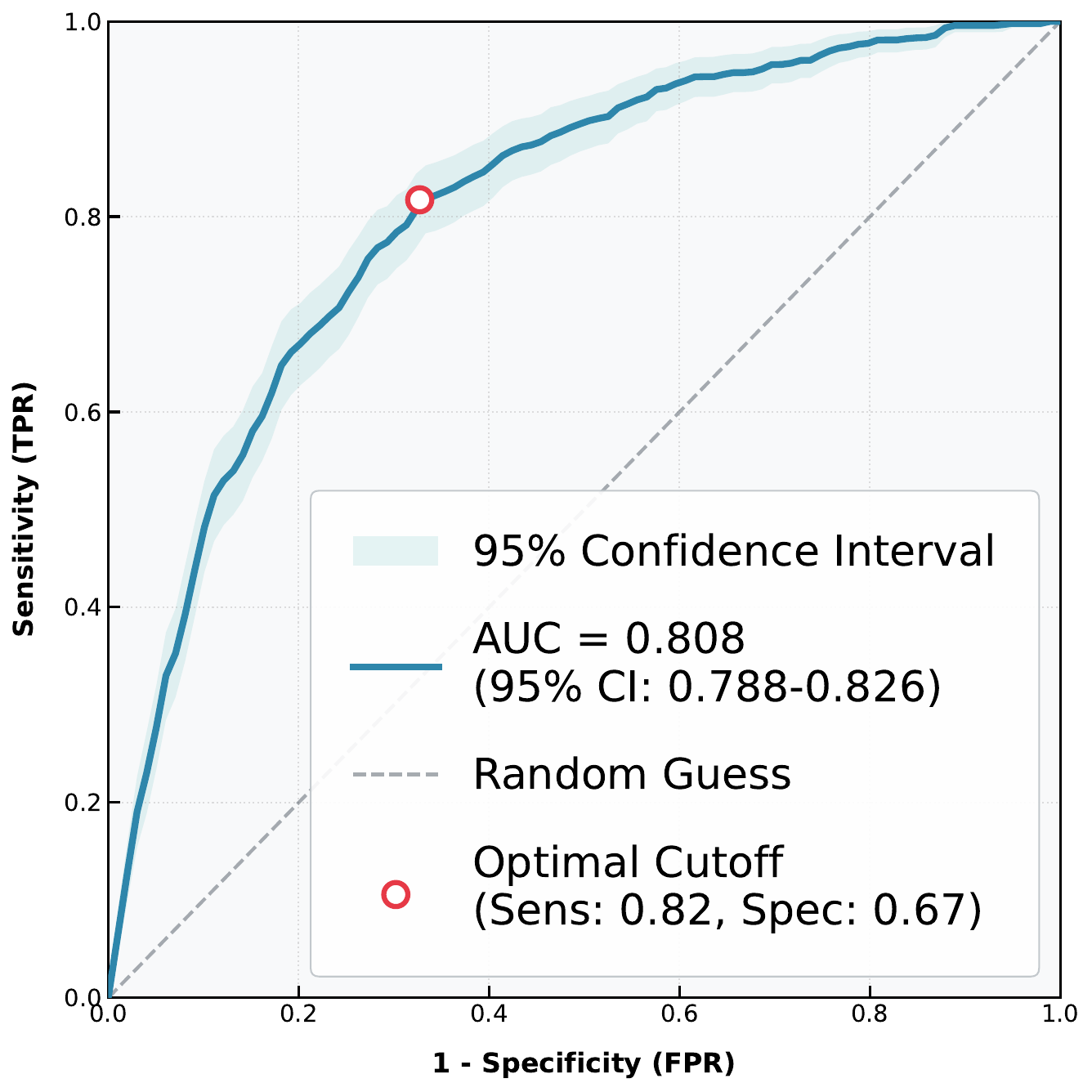}
        \caption{Independent external validation}
        \label{fig:results_roc_c}
    \end{subfigure}
    
    \caption{\textbf{ROC curves for model performance evaluation.} 
    (a) Internal testing ROC curve of PKUPH. 
    (b) Temporal validation ROC curve of PKUPH.
    (c) Independent external ROC curve of SHTMU.}
    \label{fig:overall_results}

    \vspace{8pt}
    \hrule
\end{figure}

Pocket-K showed stable discrimination across the validation sets. In the internal test set, the AUROC was 0.9364 (95\% CI 0.9135--0.9575), with a sensitivity of 83.33\% (95\% CI 78.24\%--87.43\%) and a specificity of 96.60\% (95\% CI 96.06\%--97.07\%) at the prespecified frozen threshold. In the temporal validation set, despite changes in case mix over time, the AUROC was 0.8582 (95\% CI 0.8441--0.8723), with a sensitivity of 85.05\% (95\% CI 81.84\%--87.77\%), suggesting robustness to temporal drift. In the independent external set, the AUROC was 0.8076 (95\% CI 0.7883--0.8258) despite concurrent differences in ECG platform and lower disease prevalence, as shown in Figure~\ref{fig:overall_results}.

Although discrimination decreased across increasingly heterogeneous evaluation settings, the negative predictive value remained high and exceeded 99.3\% in the external set. These findings support the role of Pocket-K as a rule-out screening tool for identifying low-risk states rather than a replacement for laboratory confirmation.

\subsection*{Pocket-K showed stronger discrimination for moderate-to-severe hyperkalemia}

\begin{figure}[htbp]
    \centering
    \captionsetup{font=small, labelfont=bf}

    \begin{subfigure}{0.32\textwidth}
        \centering
        \includegraphics[width=\linewidth]{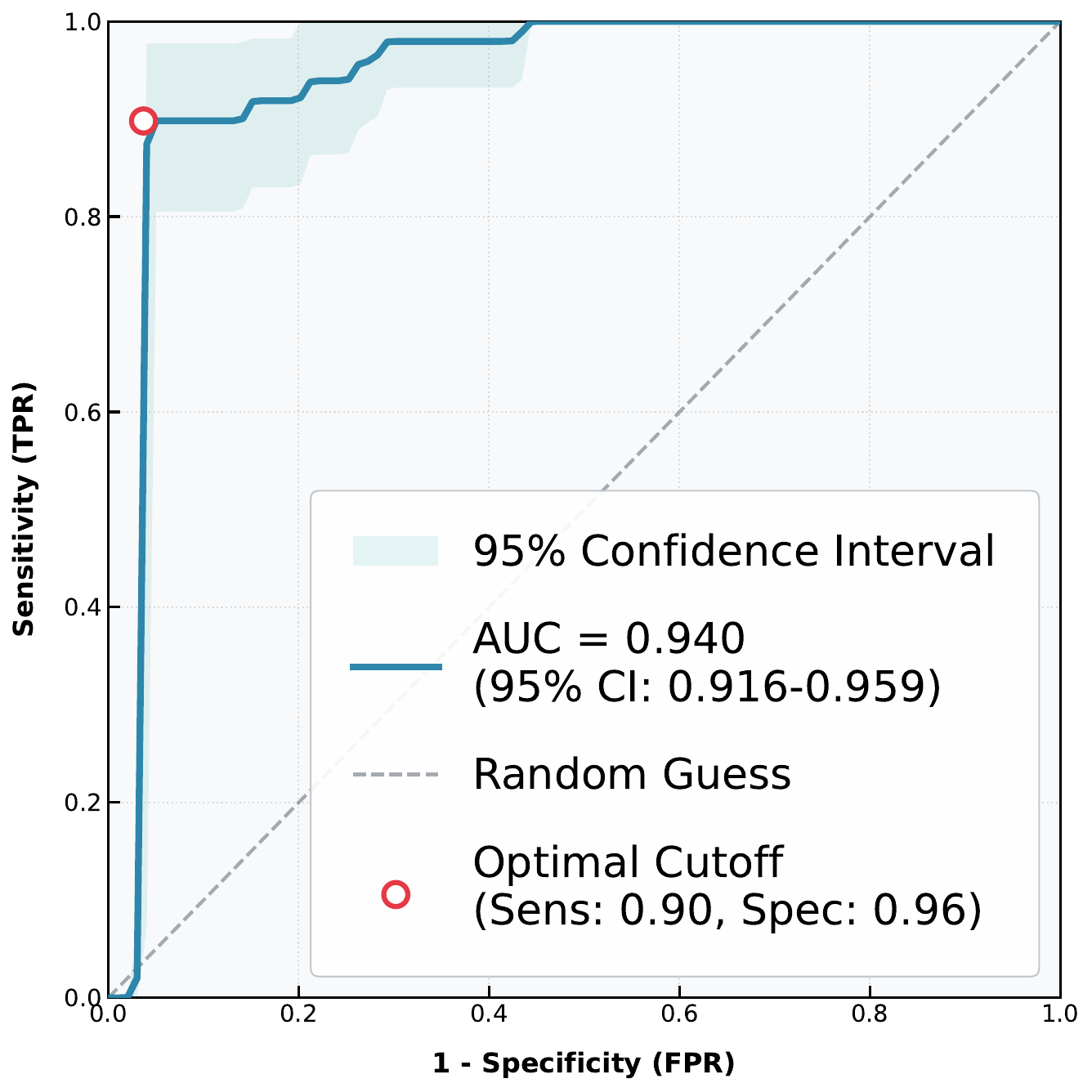}
        \caption{Temporal External Validation}
        \label{fig:severe_roc_a}
    \end{subfigure}
    \hspace{40pt}
    \begin{subfigure}{0.32\textwidth}
        \centering
        \includegraphics[width=\linewidth]{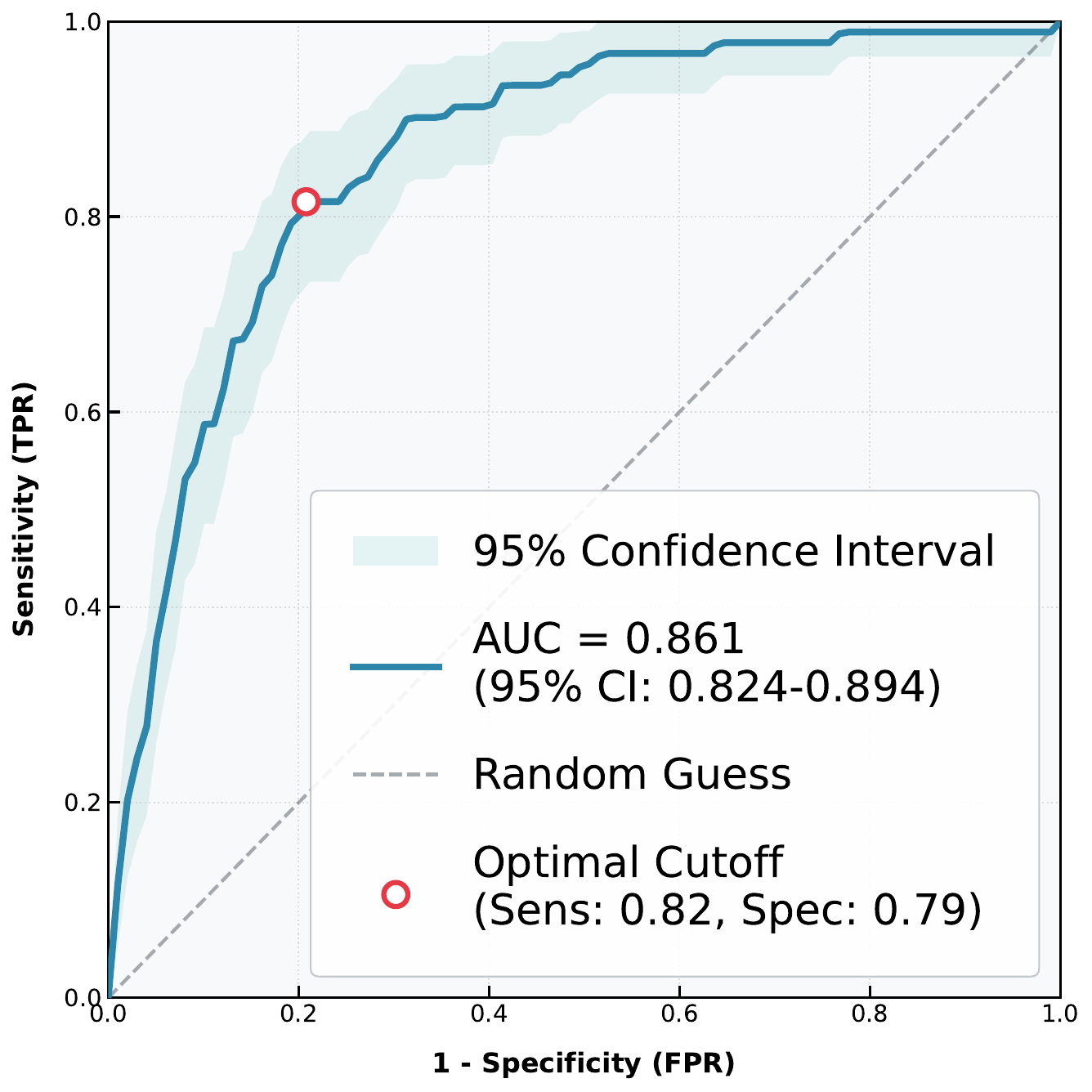}
        \caption{Independent External Validation}
        \label{fig:severe_roc_b}
    \end{subfigure}
    
    \caption{ROC curves for detection of KDIGO-defined moderate-to-severe hyperkalemia (serum potassium $\geq$ 6.0 mmol/L). (a) Temporal validation ROC curve in PKUPH. (b) Independent external validation ROC curve in SHTMU.}
    \label{fig:overall_severe_results}

    \vspace{8pt}
    \hrule
\end{figure}

Model performance was better for KDIGO-defined moderate-to-severe hyperkalemia (serum potassium $\geq$ 6.0 mmol/L). In the temporal validation set, the AUROC increased to 0.9399 (95\% CI 0.9160--0.9590), with a sensitivity of 89.80\% (95\% CI 78.24\%--95.56\%). In the independent external validation set, the AUROC was 0.8613 (95\% CI 0.8243--0.8940), with a sensitivity of 81.52\% (95\% CI 72.39\%--88.13\%), as shown in Figure~\ref{fig:overall_severe_results}.

In the independent external set, the negative predictive value for moderate-to-severe hyperkalemia was 99.91\% (95\% CI 99.86\%--99.95\%), indicating a very low probability of missed clinically urgent events in this screening setting. These findings suggest that Pocket-K was more sensitive to the more marked electrophysiological abnormalities associated with larger potassium elevations.

\subsection*{Pocket-K captured recognizable electrophysiological features of hyperkalemia}
\begin{figure}[htbp]
    \centering
    \captionsetup{font=small, labelfont=bf}

    \includegraphics[width=0.8\textwidth]{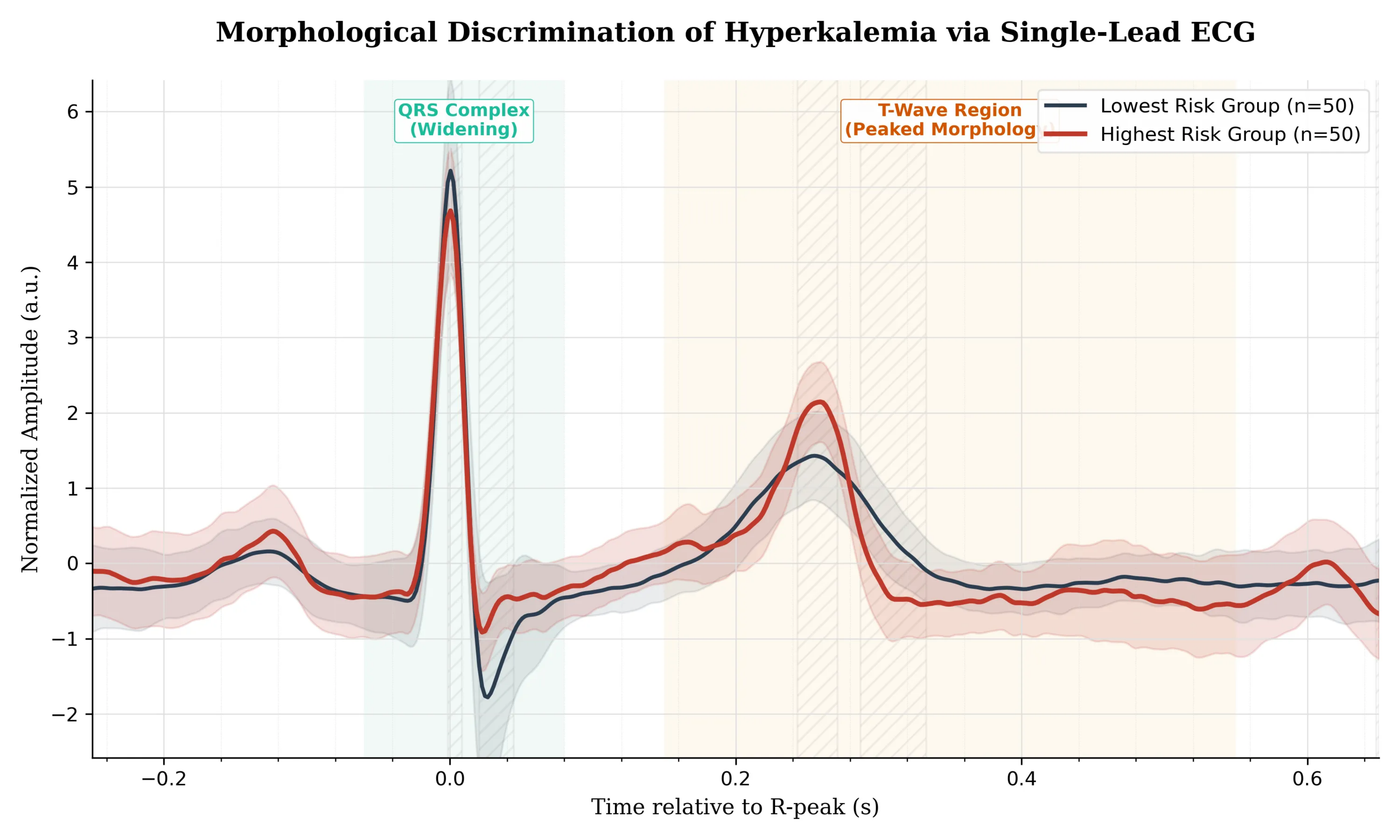}

    \caption{Signal-averaged waveform comparison. Average Lead I ECG waveforms are shown for patients correctly classified by the model as hyperkalemic (high risk, red line) and non-hyperkalemic (low risk, blue line). Shaded areas indicate standard deviation.}
    \label{fig:inter}

    \vspace{8pt}
    \hrule
\end{figure}

To examine the waveform features underlying model predictions, we stratified samples into high-risk and low-risk groups according to the predicted probability and compared the signal-averaged Lead I morphology between groups. Normalized heartbeat segments were extracted from the raw signals after temporal alignment and standardization, and group-level mean waveforms with standard deviation bands were then calculated. The clearest differences were observed in the T-wave and QRS regions, consistent with the known electrophysiological manifestations of hyperkalemia (Figure~\ref{fig:inter}). The low-risk group shows a more typical waveform pattern. These findings suggest that high-risk predictions were driven by recognizable waveform features rather than noise alone.

\subsection*{Model-predicted high risk below the hyperkalemia threshold may serve as a potential marker of greater cardiorenal burden}

\begin{figure}[htbp]
    \centering
    \captionsetup{font=small, labelfont=bf}

    \includegraphics[width=0.6\textwidth]{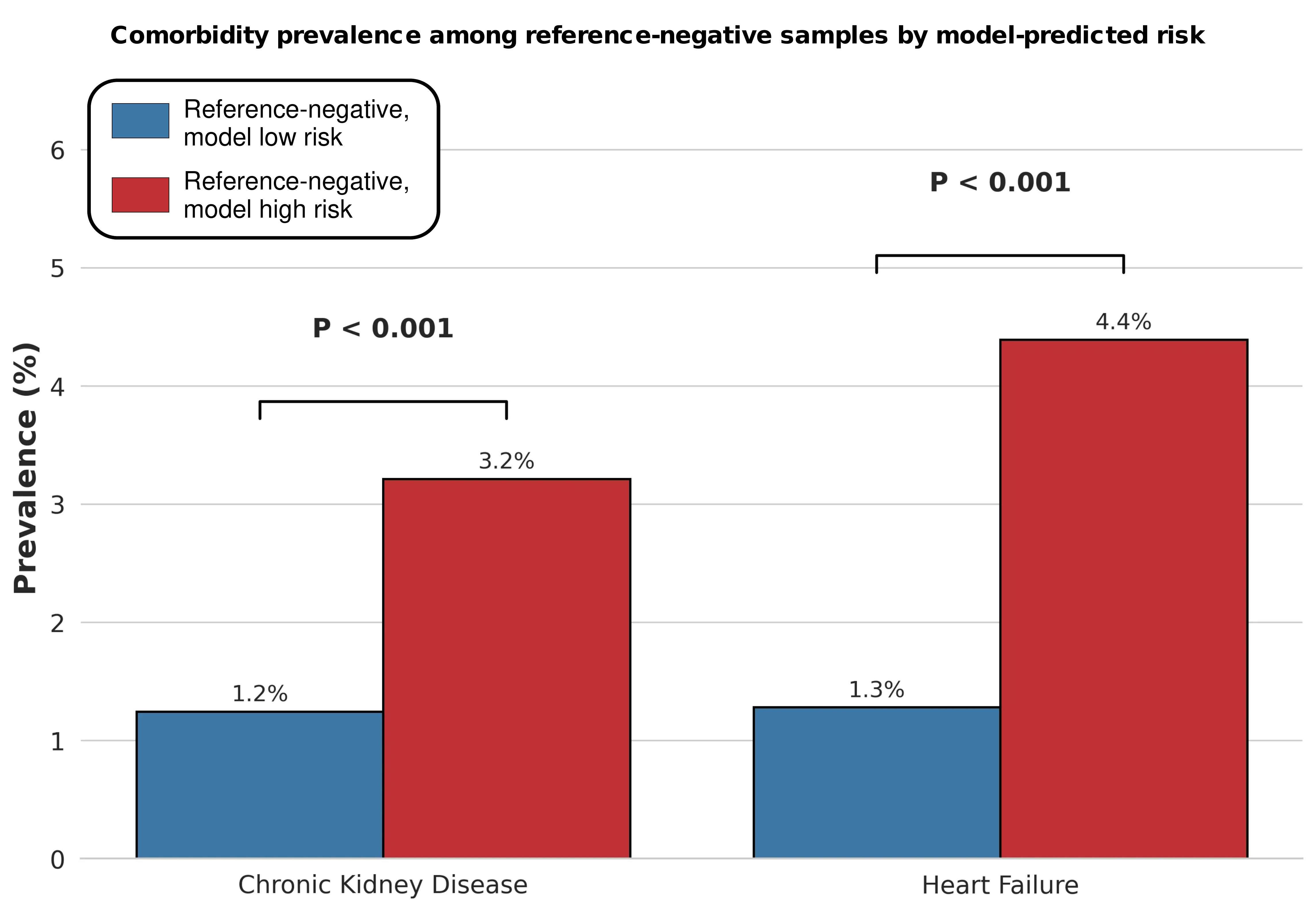}

    \caption{Clinical profiles of reference-negative samples stratified by model-predicted risk. In the external validation set, all samples shown had serum potassium values below the diagnostic threshold for hyperkalemia. They were stratified into a model low-risk group and a model high-risk group according to model output. Chronic kidney disease and heart failure were more common in the model high-risk group. The prevalence of chronic kidney disease increased from 1.2\% in the model low-risk group to 3.2\% in the model high-risk group ($P<0.001$), and the prevalence of heart failure increased from 1.3\% to 4.4\% ($P<0.001$).}
    
    \label{fig:comorbidity_fp_tn}

    \vspace{8pt}
    \hrule
\end{figure}

To better understand model behaviour among samples below the hyperkalemia threshold, we compared the clinical profiles of reference-negative samples stratified by model-predicted risk in the external validation set (Figure~\ref{fig:comorbidity_fp_tn}). Because all samples in this analysis had serum potassium values below the diagnostic threshold, the aim was to determine whether model-predicted high risk reflected random error or preferential identification of a subgroup with greater underlying disease burden. We found that chronic kidney disease and heart failure were more common in the model high-risk group than in the model low-risk group. The prevalence of chronic kidney disease increased from 1.2\% in the model low-risk group to 3.2\% in the model high-risk group ($P<0.001$), and the prevalence of heart failure increased from 1.3\% to 4.4\% ($P<0.001$). These findings suggest that model-predicted high risk below the hyperkalemia threshold was not simply random error. Instead, it may identify a subgroup with greater cardiorenal burden, even when the indexed serum potassium value does not exceed the diagnostic threshold.

\subsection*{Pocket-K tracked clinically meaningful potassium trajectories over time}

\begin{figure}[htbp]
    \centering
    \captionsetup{font=small, labelfont=bf}

    \includegraphics[width=1.0\textwidth]{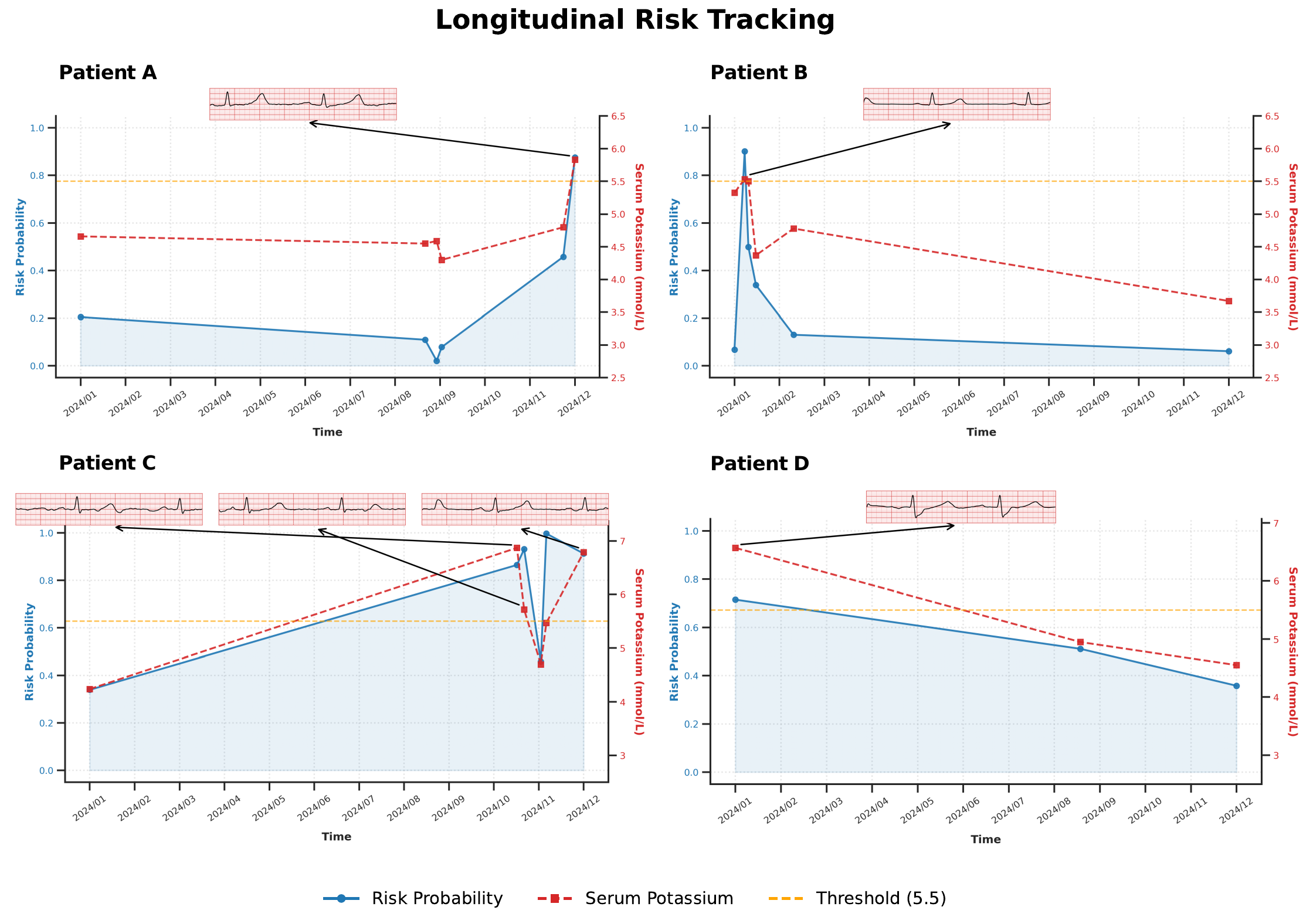}

    \caption{Longitudinal tracking of serum potassium using single-lead AI-ECG. Representative patients showing the model's ability to track within-patient potassium changes over time. The red line indicates laboratory serum potassium, and the blue dashed line indicates the model-predicted risk probability. Patient A shows progressive potassium rise with late threshold crossing. Patient B shows an early hyperkalemic episode followed by sustained recovery. Patient C shows recurrent severe fluctuation with corresponding changes in model output. Patient D shows gradual biochemical improvement accompanied by declining predicted risk. Overall, these examples suggest that model output tracked clinically meaningful potassium dynamics at the individual-patient level.}
    \label{fig:longitudinal_tracking}

    \vspace{8pt}
    \hrule
\end{figure}

To illustrate longitudinal tracking at the patient level, we selected four representative patients from the validation sets with repeated ECG--potassium measurements over time (Figure~\ref{fig:longitudinal_tracking}). These patients illustrate several clinically relevant longitudinal patterns, including progressive potassium rise, early high-risk detection with subsequent recovery, recurrent severe fluctuation, and gradual biochemical improvement after treatment.

For Patient A, serum potassium remained below the diagnostic threshold for most of follow-up, while the model assigned persistently low risk. As potassium increased late in follow-up and eventually crossed the hyperkalemia threshold, the predicted risk rose sharply in parallel. This pattern suggests that the model was sensitive to progressive potassium elevation and may be useful for identifying impending deterioration before or around threshold crossing.

For Patient B, an early episode of hyperkalemia was observed, during which the predicted risk rose to a high level. As serum potassium subsequently declined, model output also decreased and remained low during later follow-up. This pattern supports the ability of the model to detect an acute high-risk state and to return to a low-risk range after biochemical recovery.

For Patient C, potassium fluctuated markedly over time, including severe elevation, partial correction, and subsequent recurrence. Model-predicted risk changed in parallel with these biochemical shifts, remaining high during severe episodes, decreasing after interim improvement, and rising again when potassium rebounded. This pattern suggests that the model may capture recurrent disease activity during longitudinal surveillance.

For Patient D, serum potassium decreased steadily over serial measurements, and the predicted risk declined in parallel from an initially high level to a substantially lower range. This pattern suggests that model output may track treatment response or clinical recovery over time in patients with resolving hyperkalemia.

\subsection*{Pocket-K supported near-real-time inference in a handheld workflow}
\begin{figure}[htbp]
    \centering
    \captionsetup{font=small, labelfont=bf}

    \includegraphics[width=0.8\textwidth]{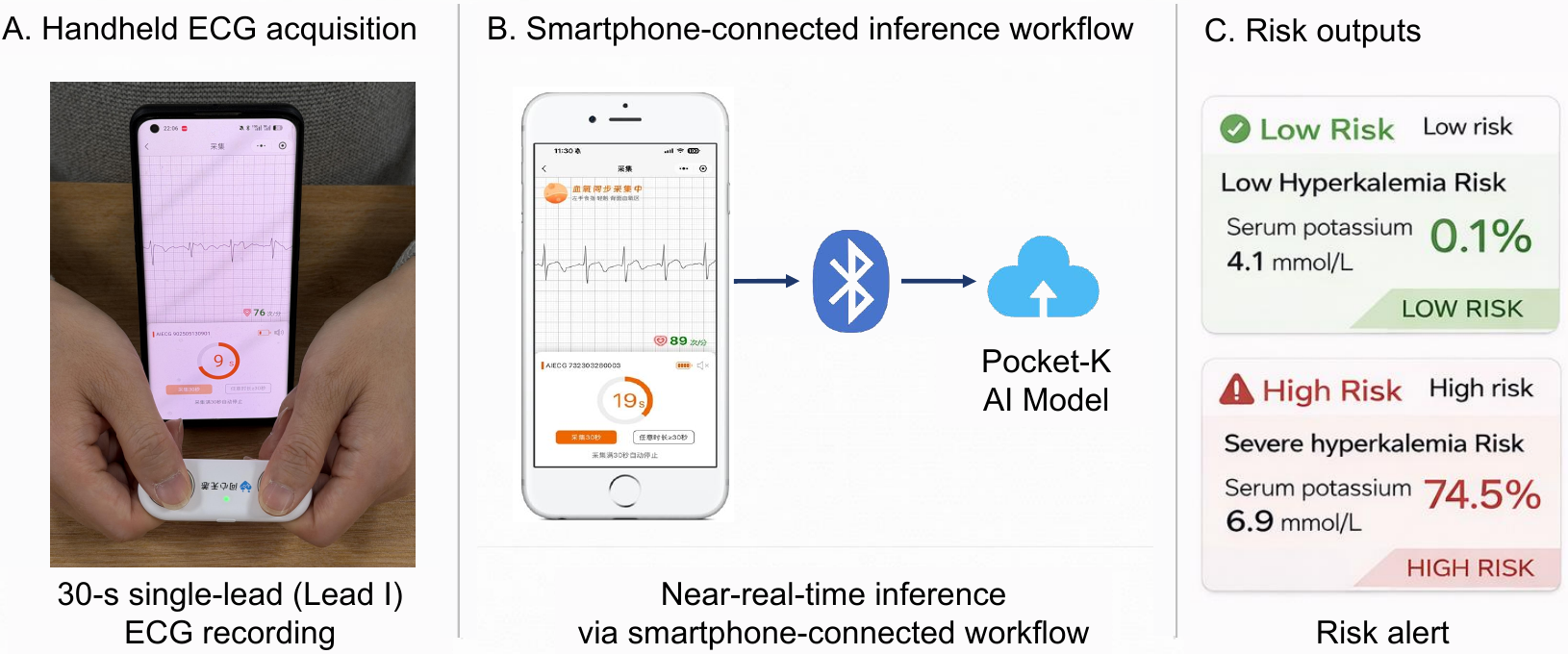}

    \caption{Proof-of-concept handheld deployment workflow of Pocket-K. (A) A handheld device paired with a smartphone application records a 30-s single-lead (Lead I) ECG. (B) The recorded signal is transmitted to the smartphone through Bluetooth and processed by a connected inference pipeline for near-real-time risk estimation. (C) Representative outputs from the prototype system show a low predicted hyperkalemia risk for a normokalemic example (serum potassium 4.1 mmol/L) and a high predicted risk for a severe hyperkalemia example (serum potassium 6.9 mmol/L).}
    
    \label{fig:fw}

    \vspace{8pt}
    \hrule
\end{figure}

To assess potential clinical use in a handheld setting, we implemented a lightweight proof-of-concept version of Pocket-K in a single-lead ECG workflow. The prototype device acquired a 30-s Lead I recording and transmitted the signal to a paired smartphone application through Bluetooth Low Energy, where near-real-time inference was performed using a connected inference pipeline (Figure~\ref{fig:fw}). In representative tests, the system returned a predicted risk of 0.1\% for a normokalemic example with a serum potassium concentration of 4.1 mmol/L and 74.5\% for a severe hyperkalemia example with a serum potassium concentration of 6.9 mmol/L within seconds. These results support the technical feasibility of rapid risk estimation in a handheld workflow.

\section*{DISCUSSION}
In this multicentre study, a single-lead AI-ECG system initialized from a pretrained foundation model showed clinically useful discrimination for hyperkalemia in internal testing, temporal validation, and independent external validation. Although performance declined as dataset heterogeneity increased, Pocket-K maintained a high negative predictive value, supporting its potential use as a non-invasive rule-out screening tool for people at increased cardiorenal risk who require repeated potassium surveillance. We also showed the technical feasibility of near-real-time inference in a handheld workflow, providing an initial basis for future prospective deployment studies.

Prior studies using 12-lead AI-ECG have reported strong discrimination for hyperkalemia and suggest that deep-learning models can detect ECG signatures that are not readily captured by routine visual interpretation \cite{lin2020deep,harmon2024mortality,chen2025monitoring}. By contrast, evidence for single-lead detection has remained limited to relatively small single-centre studies, often without rigorous external validation or evaluation across acquisition platforms \cite{attia2016novel,urtnasan2022noninvasive,chiu2024serum}. Against this background, our study extends the literature by evaluating a foundation-model-based single-lead approach in a larger multicentre setting, with temporal separation within one health system and external validation on a different ECG platform. The lower discrimination observed in external validation relative to internal testing is therefore not unexpected, but instead reflects a more realistic test of transportability for a signal-limited screening model.

These findings also help clarify the most plausible clinical role of Pocket-K. In lower-prevalence external settings, positive predictive value declined whereas negative predictive value remained high, which is consistent with the expected behaviour of a rule-out screening tool under prevalence shift. This pattern suggests that Pocket-K is better suited to repeated surveillance and triage than to definitive diagnosis. In practice, such a tool may be most useful for identifying low-risk states and for prompting confirmatory laboratory testing or clinical review in patients who require ongoing potassium surveillance, particularly those with advanced chronic kidney disease, dialysis dependence, or combined cardiac and renal disease.

The stronger discrimination observed for moderate-to-severe hyperkalemia is clinically important. In our study, this endpoint corresponded to serum potassium $\geq$ 6.0 mmol/L and therefore spanned KDIGO-defined moderate and severe hyperkalemia. This finding is biologically plausible, because larger potassium elevations are more likely to produce overt repolarization and conduction abnormalities, increasing signal detectability on ECG. From a clinical perspective, improved performance in this range is especially relevant, as the main priority of a screening tool in high-risk populations is to minimise missed urgent events rather than to replace biochemical confirmation.

The interpretability and phenotype analyses further support the biological plausibility of the model. Group-level waveform differences were concentrated in the T-wave and QRS regions, consistent with established electrophysiological manifestations of hyperkalemia. In addition, among samples below the diagnostic threshold, model-predicted high risk was more common in patients with chronic kidney disease and heart failure. This observation raises the possibility that model outputs may reflect broader cardiorenal electrophysiological stress rather than potassium concentration alone at a single indexed time point. Although this interpretation remains hypothesis-generating, it may help explain why some apparently discordant predictions are not clinically trivial.

The longitudinal case analyses and handheld prototype provide complementary evidence for potential clinical translation. Serial examples suggested that model-predicted risk changed in parallel with within-patient potassium trajectories, supporting the possibility that Pocket-K may capture evolving potassium-related electrophysiology over time. The proof-of-concept handheld workflow further showed that near-real-time inference is technically feasible in a connected smartphone pathway. 

Our study also provides initial evidence of translational feasibility. A handheld single-lead ECG workflow supported near-real-time inference through a connected smartphone-based pathway and may allow immediate feedback without additional invasive burden. However, this part of the study was designed as a proof of concept rather than an evaluation of clinical effectiveness. Prospective studies using native handheld and wearable recordings are therefore needed to assess usability, signal-quality failure modes, calibration, and the effect of AI-guided screening on repeat testing, clinical decision-making, and health-care use.

This study has several limitations. First, it used historical routinely collected clinical data and therefore cannot establish whether model-guided screening improves clinical outcomes or workflow efficiency. Second, although we included temporal and independent external validation, both health systems were located in China, and broader validation across populations, geographic settings, and device form factors is still needed. Third, the single-lead input was derived from clinical 12-lead ECG systems rather than trained and validated primarily on native consumer-device recordings. Finally, model performance in the presence of rhythm abnormalities, motion artefact, and low-quality recordings requires more systematic evaluation before large-scale real-world deployment.


In conclusion, Pocket-K provides multicentre evidence that a foundation-model-based single-lead AI-ECG approach can support hyperkalemia screening across temporally separated and externally independent validation settings. Its high negative predictive value suggests potential value as a scalable rule-out tool for frequent potassium surveillance in high-risk populations. Prospective studies should now assess how such systems can be integrated into handheld and wearable care pathways to support earlier detection while preserving safety and clinical interpretability.

\section*{METHODS}

\subsection*{Study design and data sources}

This multicenter observational study used routinely collected de-identified clinical ECG and laboratory data and adhered to the TRIPOD+AI reporting guidelines for clinical artificial intelligence \cite{collins2024tripod+}. Data were collected from two geographically independent tertiary health systems in China: Peking University People's Hospital (PKUPH, Beijing) and The Second Hospital of Tianjin Medical University (SHTMU, Tianjin). The PKUPH set was extracted from the MedEx ECG management system (January 2016--November 2024), while the SHTMU set was retrieved from the Nalong system (January 2024--December 2024). The study was approved by the Biomedical Ethics Committee of Peking University (IRB00001052-23152), and the requirement for informed consent was waived because the analysis used de-identified routinely collected clinical data.

\subsection*{Study population assembly and data partitioning}

\begin{figure}[htbp]
    \centering
    \captionsetup{font=small, labelfont=bf}

    \includegraphics[width=0.95\textwidth]{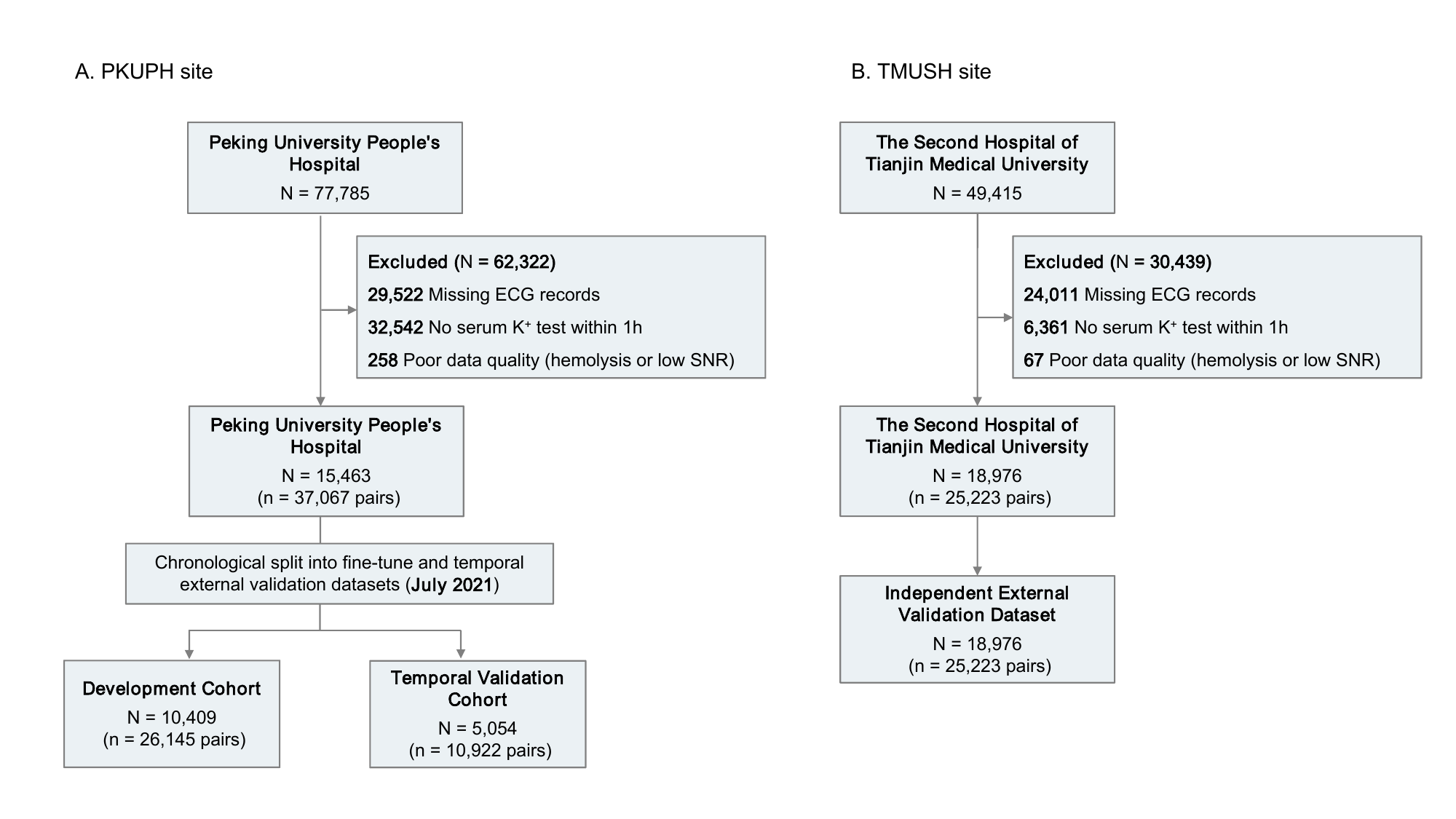}

    \vspace{4pt} 

    \caption[STARD flowchart of study population selection.]{%
    \textbf{STARD flowchart for study population selection and multicenter dataset partitioning.}
    Panel \textbf{A} shows set construction at PKUPH, where 77,785 patients were initially screened. A total of 62,322 patients were excluded through a multistage quality-control pipeline: 29,522 had no ECG records, 32,542 had no eligible venous serum potassium measurement within a $\pm$1-hour window for ECG-anchored pairing, and 258 were removed because of poor data quality, including confirmed laboratory haemolysis or low signal-to-noise ratio. The remaining PKUPH population ($N=15,463$; $n=37,067$ pairs) was partitioned chronologically at July 2021 into a development set ($N=10,409$; $n=26,145$ pairs) and a temporal validation set ($N=5,054$; $n=10,922$ pairs). 
    Panel \textbf{B} shows independent external validation at SHTMU. Of 49,415 initially screened patients, 30,439 were excluded using the same criteria (missing ECGs: $n=24,011$; no eligible potassium measurement within $\pm$1 hour for ECG-anchored pairing: $n=6,361$; poor data quality: $n=67$), resulting in a final independent external validation set of 18,976 patients with 25,223 ECG--$K^+$ pairs.
    In all panels, $N$ denotes the number of unique patients and $n$ denotes the number of ECG--$K^+$ paired records.}

    \label{fig:cohort1}

    \vspace{4pt}
    \hrule
\end{figure}

As illustrated in Figure~\ref{fig:cohort1}, we first screened all patients with available ECG data during the study periods. After exclusion of haemolysed samples, venous serum potassium measurements were linked to ECG recordings using an ECG-anchored pairing strategy. Specifically, for each ECG, we searched for potassium measurements obtained within $\pm$1 h and retained the closest eligible measurement to form a unique ECG--potassium pair. ECGs without any eligible potassium measurement within the predefined time window were excluded.

The PKUPH population was partitioned chronologically at July 2021. Records before this cutoff formed the development set, and records after this cutoff formed the temporal validation set. To avoid information leakage, we imposed strict patient-level non-overlap: any patient appearing in the development period was excluded entirely from the temporal validation set, even if later records were available. Within the PKUPH development set, unique patients were further divided at the patient level into a fine-tune set, a model-selection set, and an internal test set in an 8:1:1 ratio according to a prespecified protocol. The geographically independent SHTMU dataset was reserved exclusively for external validation.

Because individual patients could contribute multiple ECG--potassium pairs across time, the number of unique patients ($N$) differed from the number of paired records ($n$).

\subsection*{Definition of clinical comorbidities}

Chronic kidney disease and heart failure were defined a priori using structured diagnosis information recorded on or before the index ECG. In the PKUPH sets, diagnoses were obtained from the hospital diagnosis table (field: diagnosis name), and in the SHTMU set, from the structured diagnosis string field (\texttt{clinic\_diag\_str}). Because the original diagnosis entries were recorded in Chinese, comorbidities were identified using prespecified rule-based keyword searches applied to the original diagnosis strings. The English terms reported below describe the diagnostic concepts used for classification rather than the literal source strings.

Chronic kidney disease was defined using diagnosis strings corresponding to chronic kidney disease or chronic kidney failure in the chronic setting, including chronic kidney disease, chronic renal insufficiency, chronic renal failure, end-stage kidney disease, end-stage renal disease, uraemia, and CKD-equivalent diagnostic expressions. Isolated non-specific terms such as renal insufficiency or renal failure were not used unless they were explicitly documented as chronic.

Heart failure was defined using diagnosis strings corresponding to heart failure, including heart failure, congestive heart failure, left heart failure, right heart failure, biventricular heart failure, heart failure with preserved ejection fraction, and heart failure with reduced ejection fraction.

These comorbidity variables were used for baseline set description and for the analysis of reference-negative samples stratified by model-predicted risk in the external validation set.

\subsection*{Reference standard and single-lead signal preprocessing}

Venous serum potassium served as the reference standard. To improve physiological correspondence between ECG waveforms and laboratory measurements, each ECG was paired with the nearest eligible potassium result within a $\pm$1 h window after exclusion of haemolysed samples. This ECG-anchored pairing strategy was designed to reduce temporal mismatch between the model input and the biochemical reference label.

Lead I was extracted from 12-lead ECG recordings to construct a single-lead input. All signals followed a unified preprocessing pipeline. We applied a 0.5--40 Hz band-pass filter to reduce baseline drift, powerline interference, and myoelectric noise, segmented the waveform into non-overlapping 10-s clips, and resampled each clip to 500 Hz by linear interpolation to match the input specification of the pretrained foundation model. Each clip was then z-score normalised to reduce amplitude-scale differences across acquisition systems.

\subsection*{Model development}
Pocket-K was developed by fine-tuning ECGFounder, a pretrained ECG foundation model. ECGFounder adopts a Net1D-style architecture with stage-wise scaling inspired by RegNet and has been pretrained on more than ten million ECG recordings collected across multiple countries \cite{li2025electrocardiogram, radosavovic2020designing, hong2020holmes}. This large-scale pretraining enables the model to capture general representations of cardiac depolarisation and repolarisation and reduces downstream dependence on sample size and data homogeneity. All fine-tuning experiments were initialised from publicly available ECGFounder pretrained weights.

The model was fine-tuned using binary cross-entropy loss and the Adam optimiser with an initial learning rate of $1\times10^{-4}$. Training was performed for up to 30 epochs. Model selection was based on AUROC in the model-selection dataset. If validation AUROC did not improve for 10 consecutive epochs, the learning rate was reduced by a factor of 0.1. The checkpoint with the best validation AUROC was retained for subsequent evaluation on the internal test dataset, temporal validation dataset, and external validation sets.

\subsection*{Performance evaluation and statistical analyses}
The primary endpoint was discrimination for hyperkalemia, defined as venous serum potassium greater than 5.5 mmol/L. A prespecified secondary endpoint was discrimination for moderate-to-severe hyperkalemia (serum potassium $\geq$ 6.0 mmol/L). According to the KDIGO 2024 CKD guideline, moderate hyperkalemia is defined as 6.0--6.4 mmol/L and severe hyperkalemia as $\geq$6.5 mmol/L\cite{kidgo}. Model discrimination was primarily assessed using the area under the receiver operating characteristic curve (AUROC).

Because individual patients could contribute multiple ECG--potassium pairs, statistical uncertainty was quantified using clustered bootstrap resampling at the patient level with 2000 resamples. This framework was used to estimate 95\% confidence intervals for AUROC and threshold-dependent performance metrics and to compare model performance across internal test, temporal validation, and external validation sets without treating repeated paired records from the same patient as independent observations.

\subsection*{Additional analyses and explainability}

We performed additional analyses to evaluate clinically relevant model behaviour beyond binary discrimination. First, moderate-to-severe hyperkalemia was analysed as a high-priority subgroup because of its greater clinical urgency. Second, waveform-level explainability analysis was performed by stratifying samples according to model-predicted risk and comparing signal-averaged heartbeat morphology between high-risk and low-risk groups, with particular attention to regions spanning the T wave and QRS complex. We also compared clinical phenotypes between false-positive and true-negative samples in the external validation set to determine whether false-positive predictions were enriched for cardiorenal comorbidity.

To assess whether model outputs tracked disease dynamics at the individual level, we examined representative longitudinal patients with repeated ECG--potassium pairs over time. Finally, for proof-of-concept deployment, we evaluated inference latency and qualitative risk prompting in a handheld single-lead workflow.

\subsection*{Proof-of-concept handheld deployment}
For proof-of-concept deployment, the handheld device acquired a 30-s single-lead ECG recording. To maintain consistency with the temporal scale used during model development, each recording was divided into three consecutive 10-s clips for inference. Clip-level outputs were then aggregated to generate a measurement-level risk probability. This design preserved temporal-scale consistency between model development and prototype deployment.

\section*{Declaration statements}

\subsection*{Data Availability}
The data that support the findings of this study are not publicly available due to restrictions imposed by institutional ethics committees and data governance policies of the participating hospitals. Access to the data may be considered upon reasonable request to the corresponding author, subject to approval by the relevant ethics committees.

\subsection*{Code Availability}
The code used for model development and evaluation in this study is publicly available at \url{https://github.com/Tangoz1003/Pocket-K}.

\subsection*{Acknowledgements} 
Shenda Hong is supported by the National Natural Science Foundation of China (62102008), CCF-Tencent Rhino-Bird Open Research Fund (CCF-Tencent RAGR20250108), CCF-Zhipu Large Model Innovation Fund (CCF-Zhipu202414), PKU-OPPO Fund (BO202301, BO202503), Research Project of Peking University in the State Key Laboratory of Vascular Homeostasis and Remodeling (2025-SKLVHR-YCTS-02), and the Beijing Municipal Science and Technology Commission (Z251100000725008).

Luxia Zhang is supported by the National Natural Science Foundation of China (72125009) and grant from the Noncommunicable Chronic Diseases-National Science and Technology Major Project of China (No. 2025ZD0547500).

Kangyin Chen is supported by the National Natural Science Foundation of China (82470527) and the Key Science and Technology Support Project of Tianjin Science and Technology Bureau (24ZXGZSY00130).

The authors thank all collaborators and participating institutions for their support and contributions to this research.

\subsection*{Author Contributions}
Gongzheng Tang, Qinghao Zhao, and Guangkun Nie contributed equally to this work. Gongzheng Tang contributed to methodology development, model implementation, validation, formal analysis, and drafting of the manuscript. Qinghao Zhao contributed to result interpretation and manuscript revision. Guangkun Nie contributed to data acquisition, preprocessing, and curation. Yujie Xiao, Shijia Geng, Donglin Xie, Shun Huang, Deyun Zhang, Xingchen Yao, and Jinwei Wang contributed to data preparation and investigation. Kangyin Chen, Luxia Zhang, and Shenda Hong conceived and supervised the study, provided resources, guided study design and interpretation, and revised the manuscript. All authors reviewed and approved the final manuscript. Kangyin Chen, Luxia Zhang, and Shenda Hong are corresponding authors and take responsibility for the integrity of the work.

\subsection*{Competing Interests}
Shenda Hong is an Associate Editor of \textit{npj Digital Medicine}. Shenda Hong was not involved in the journal’s review of, or decisions related to, this manuscript. The other authors declare no competing financial or non-financial interests related to this study.

\bibliography{reference}

\bigskip

\end{document}